\documentclass[conference]{IEEEtran}
\IEEEoverridecommandlockouts
\usepackage{cite}
\usepackage{amsmath,amssymb,amsfonts}
\usepackage{algorithmic}
\usepackage{graphicx}
\usepackage{textcomp}
\usepackage{xcolor}
\usepackage{multirow}
\def\BibTeX{{\rm B\kern-.05em{\sc i\kern-.025em b}\kern-.08em
    T\kern-.1667em\lower.7ex\hbox{E}\kern-.125emX}}
\begin{document}

\title{Enhancing Relation Extraction Using Syntactic Indicators and Sentential Contexts\\
}

\author{
\IEEEauthorblockN{1\textsuperscript{st} Qiongxing Tao}
\IEEEauthorblockA{\textit{School of Computer Engineering}
\\ \textit{ and Science} \\
\textit{Shanghai University}\\
Shanghai, China \\
taoqiongxing@shu.edu.cn}
\and
\IEEEauthorblockN{2\textsuperscript{nd} Xiangfeng Luo}
\IEEEauthorblockA{\textit{School of Computer Engineering}
\\ \textit{ and Science} \\
\textit{Shanghai University}\\
Shanghai, China \\
Luoxf@shu.edu.cn}
\and
\IEEEauthorblockN{3\textsuperscript{rd}  Hao Wang}
\IEEEauthorblockA{\textit{School of Computer Engineering}
\\ \textit{ and Science} \\
\textit{Shanghai University}\\
Shanghai, China \\
wang-hao@shu.edu.cn}
}

\maketitle

\begin{abstract}
State-of-the-art methods for relation extraction consider the sentential context by modeling the entire sentence. However, syntactic indicators, certain phrases or words like prepositions that are more informative than other words and may be beneficial for identifying semantic relations. Other approaches using fixed text triggers capture such information but ignore the lexical diversity. To leverage both syntactic indicators and sentential contexts, we propose an indicator-aware approach for relation extraction. Firstly, we extract syntactic indicators under the guidance of syntactic knowledge. Then we construct a neural network to incorporate both syntactic indicators and the entire sentences into better relation representations. By this way, the proposed model alleviates the impact of noisy information from entire sentences and breaks the limit of text triggers. Experiments on the SemEval-2010 Task 8 benchmark dataset show that our model significantly outperforms the state-of-the-art methods.
\end{abstract}


\section{Introduction}

Relation extraction is the task of assigning a semantic relation to the target entity pair in a given sentence. Accurately extracting semantic relations from unstructured texts is important for many natural language applications, such as information extraction \cite{fader2011identifying}\cite{wu2010open}, question answering \cite{fan2005using}\cite{yih2014semantic}, and construction of semantic networks \cite{miller1991semantic}\cite{vossen1998multilingual}. 

Recent approaches for relation extraction primarily concentrate on deep neural networks \cite{zeng2014relation, dos2015classifying, zhang2015bidirectional, zhou2016attention, xiao2016semantic, huang2016attention}. Commonly, these models encode the entire sentence to capture the contextual information for relation representation, based on the assumption that each word in a sentence helps classify relations. A majority of these methods use entity information to improve the performance of relation extraction, such as entity position \cite{zeng2014relation}\cite{dos2015classifying}, entity hypernym \cite{zeng2014relation} and latent entity typing \cite{lee2019semantic}. They all assume that the information related to target entities is more important. However, these models have two disadvantages: 
first, some words in a sentence irrelevant to the relation are as noises to classification;
second, entity information is very limited in predicting relation types and the contributions from other words are prone to be ignored. 

Besides, a few approaches rely on particular lexical constraints \cite{fader2014open} and relation triggers \cite{bjorne2011extracting} that explicitly indicate the occurrence of relations in sentences. However, these methods are not suited to cases where no relation trigger found in the sentence.

\begin{figure}[t]
\centerline{\includegraphics[width=1\linewidth]{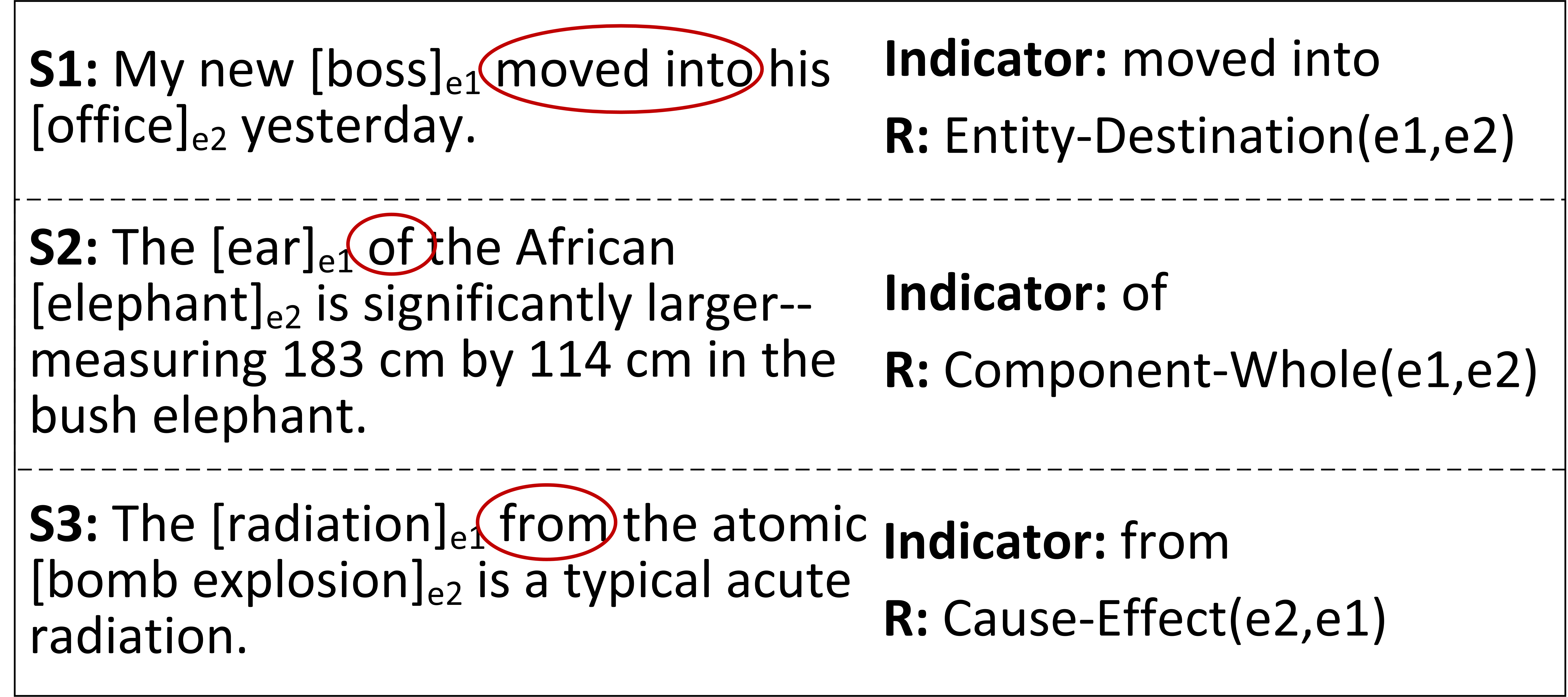}}
\caption{The decisive influence of syntactic indicators in identifying relations. The right part shows the syntactic indicator and the correct relation between target entities for each instance.}
\label{fig:1example}
\end{figure}

In this paper, we revisit the problem from another perspective. As shown in Fig.~\ref{fig:1example}, the phrase “moved into” is the key to identify the relation type {\em Entity-Destination(e1,e2)}. On the contrary, it is insufficient to recognize relation types by the linguistic features about entity “boss” and entity “office”, let alone a non-existent explicit relation trigger. Intuitively, words like “of” and “from” are informative for relation extraction. Here and after we call this kind of words or phrases syntactic indicator. Syntactic indicator contains vibrant information for identifying semantic relations between target entities. Besides, the words “My, new, yesterday” in the first sentence are ubiquitous while not useful for relation identification. We can acquire better performance by reducing their impact. 

Therefore, we propose an indicator-aware neural model to condition both the syntactic indicator and the sentential context for better performance on relation extraction. This is achieved by a two-phase process. Firstly, under the guidance of syntactic knowledge, we extract syntactic indicators by removing unrelated words through entity disambiguation, principal component extraction, and unrelated entity removal. Then, we feed both of entire sentences and syntactic indicators into a contextual encoder based on the pre-trained BERT (Bidirectional Encoder Representations from Transformers) \cite{devlin2019bert} to encode the semantic relation representations. The syntactic indicator is treated as the principal constraint on the contextual representation. By this way, the proposed model takes advantage of the relevant information and reduces the impact of noisy words. Our main contributions are listed as follows:
\begin{itemize}
\item We define syntactic indicators that help to distinguish relation types and extract syntactic indicators under the guidance of syntactic knowledge, which is conducive to capture the important information and reduce noisy information that is irrelevant to relation extraction.
\item We propose an indicator-aware neural model using the pre-extracted indicators to improve relation extraction, which makes use of the key information by imposing constraints on contextual representations for better prediction.
\item The proposed model obtains an F1-score of 90.36\% on the benchmark dataset, outperforming the state-of-the-art methods. More ablation experiments demonstrate that incorporating syntactic indicators into contextual representations significantly improves the performance of relation extraction.
\end{itemize}

\section{Related Work}
Conventional non-neural models for relation extraction include feature-based models \cite{kambhatla2004combining}\cite{suchanek2006combining} and kernel-based models \cite{qian2008exploiting}\cite{mooney2006subsequence}. These methods invariably suffer from error propagation due to their high dependence on the manual feature extraction process. Besides they may omit useful information for relation extraction. Therefore, the performance of these methods is very limited.

Recently, a variety of works for relation extraction focus on deep neural networks. These methods mitigate the problem of error propagation and show promising results. On the one hand, Zeng et al. \cite{zeng2014relation} propose a deep convolutional neural network (CNN) to address this task. They utilize sentence-level features and lexical level features, including entities, left and right tokens of entities and WordNet hypernyms of entities. Santos et al. \cite{dos2015classifying} propose the Ranking CNN (CR-CNN) model using a new rank loss to reduce the impact of artificial classes. They also demonstrate that the words between target nominals are almost as useful as using positing embeddings. Inspired by their work, we extract syntactic indicators from the text between two entities. Shen and Huang \cite{huang2016attention} propose attention-based convolutional neural network (Attention-CNN), which employs a word-level attention mechanism to get the critical information for relation representation. These methods have limitations on learning sequence structures because of the shortages of convolutional neural networks. On the other hand, the RNN-based models show outstanding performance in learning the linguistic structure in text. Zhang and Wang \cite{zhang2015relation} propose a bidirectional recurrent neural network (Bi-RNN) to learn the long-term dependency between two entities, however, it suffers the vanishing gradient problem in RNNs. Soon after, Zhang et al. \cite{zhang2015bidirectional} apply the bidirectional LSTM network (Bi-LSTM) and utilize the word position and external features to improve the performance of relation extraction, including POS tags, named entity information, and dependency parse. In \cite{zhou2016attention}, Zhou et al. apply attention mechanisms in bidirectional LSTM networks (Attention Bi-LSTM). Xiao and Liu \cite{xiao2016semantic} separate each sentence into three context subsequences according to the locations of two target entities and use a Hierarchical Recurrent Neural Network with two Attention Bi-LSTM networks (Hier Attention Bi-LSTM) to get a better result. Most recently, Lee et al. \cite{lee2019semantic} propose a model incorporating entity-aware attention mechanisms with a latent entity typing (LET) and obtain state-of-the-art performance. 

Approaches mentioned above encode the entire sentence to capture the contextual features, resulting in ignorance of other important features in sentences. Although a number of methods utilize various entity information, including the entity position, entity semantics, latent entity typing, and entity hypernyms, and such information holds an irreplaceable impact on identifying relations, it is too limited to fully capture distinctive features.

There are also some works concentrate on relation triggers, the phrases that explicitly expresses the occurrence of one relationship in the given text. Bj{\"o}rne et al. \cite{bjorne2011extracting} propose the relation triggers and determine their arguments to reduce the complexity of the task. Open IE systems ReVerb \cite{fader2014open} also uses special phrases to identify different relation types by lexical constraints. Nevertheless, there are many texts with no explicit relation trigger inside, semantical relations cannot be extracted from such sentences with these methods. Unlike these methods, our approach makes use of syntactic indicators, which can be able to vary with the different expressions of semantic relations rather than match fixed phrases templates.

\begin{figure*}
\centerline{\includegraphics[width=1\linewidth]{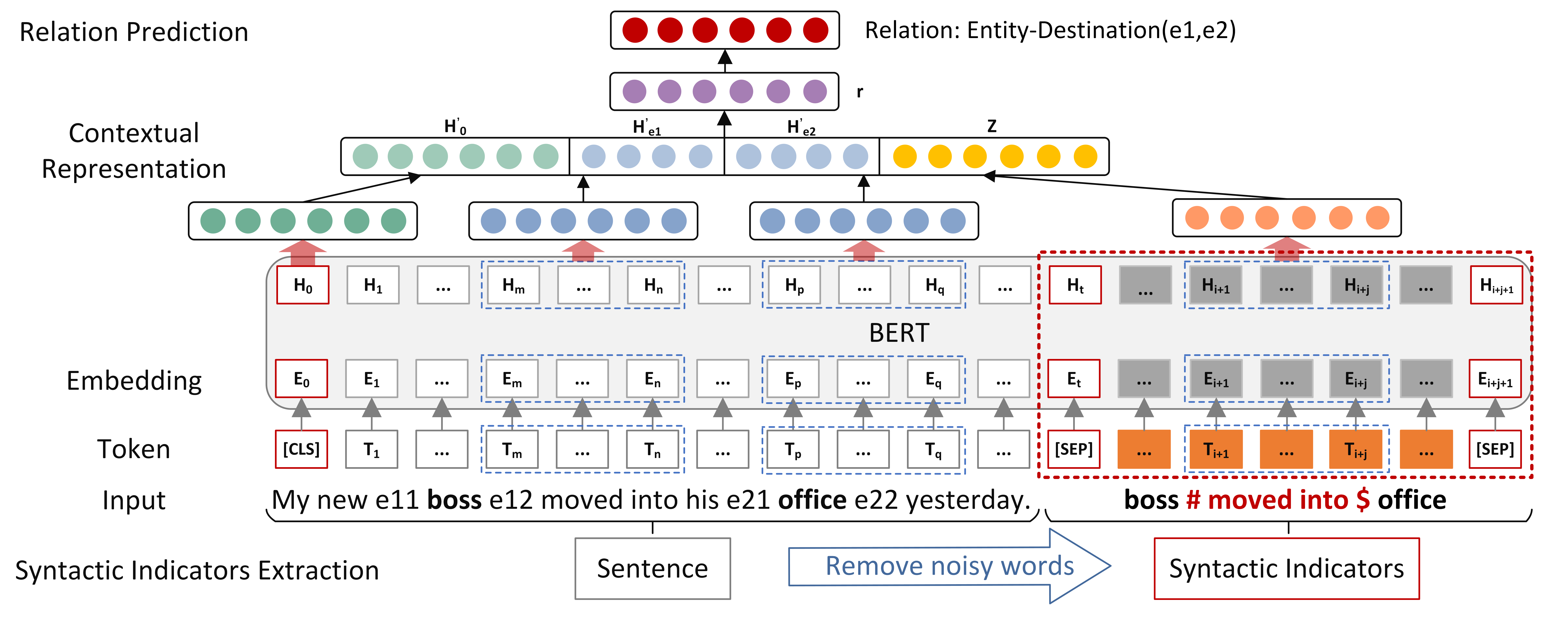}}
\caption{The overall architecture of the proposed model.}
\label{fig:architecture}
\end{figure*}

Pre-trained Language models have shown the great success on many NLP tasks \cite{dai2015semi}\cite{howard2018universal}. Especially, BERT proposed by Devlin et al. shows a significant impact \cite{devlin2019bert}, which learns the deep bidirectional representations by jointly conditioning on both left and right context in the training procedure. It has been applied to multiple NLP tasks and obtains new start-of-the-art results on eleven tasks, such as text classification, sequence labeling, and question answering. In recent research, Wu and He \cite{wu2019enriching} propose an R-BERT model, which employs the pre-trained BERT language model and reaches the top of the leaderboard in relation extraction.

By the way, related works on the relation extraction can be mainly grouped into two categories, supervised methods \cite{zeng2014relation}\cite{xiao2016semantic}\cite{socher2012semantic} and distant supervised methods \cite{mintz2009distant, zeng2015distant, min2013distant}. They are different in whether the data contains a large number of noisy labels. Supervised methods without noisy labels achieve more reliable results, which play a dominant role in the relation classification. In this paper, we focus on supervised relation extraction.

\section{Our Model}

In this section, we first give an overview of the proposed indicator-aware neural model. After that, we present each module in details.

\subsection{Model Architecture}

The overall architecture of the proposed model is shown in Fig.~\ref{fig:architecture}. 

Given a sentence, we first extract the corresponding syntactic indicator under the guidance of syntactic knowledge (the process detailed in the following paragraphs). Subsequently, the entire sentence and the indicator sequence are concatenated after WordPiece tokenization \cite{sennrich2016neural}. Then, we feed the aggregate token sequence into a BERT-based contextual encoder to learn the deep bidirectional representation for each token. The final representations of the aggregate sequence, two entities, and the syntactic indicators are respectively acquired with different operations in the later network layers. At last, these vector representations are concatenated to produce a final prediction distribution.

\subsection{Definition of Syntactic Indicators}

{\bf \em Definition: }The syntactic indicator is certain words or phrases in a sentence, providing essential information to identify the semantic relation between target entities. 

Different from text triggers, syntactic indicators are rich in manifestation rather than match fixed phrase templates. Each sentence produces an exclusive syntactic indicator. It may consist of any verbs, prepositions, pronouns or phrases, relying on the current language expression. As shown in Fig.~\ref{fig:3examples}, “caused by” is the syntactic indicator in the first instance. Accordingly, we can affirm that relation {\em Cause-Effect(e2,e1)} exists in two target enties \begin{math}e_1\end{math}={\em shock} and \begin{math}e_2\end{math}={\em attack}. Similarly in the other two instances, we can recognize relation {\em Content-Container(e1,e2)} and relation {\em Instrument-Agency(e2,e1)} based “are enclosed in” and “using”, respectively.

\subsection{Syntactic Indicator Extraction}

\begin{figure}[ht]
\centerline{\includegraphics[width=1\linewidth]{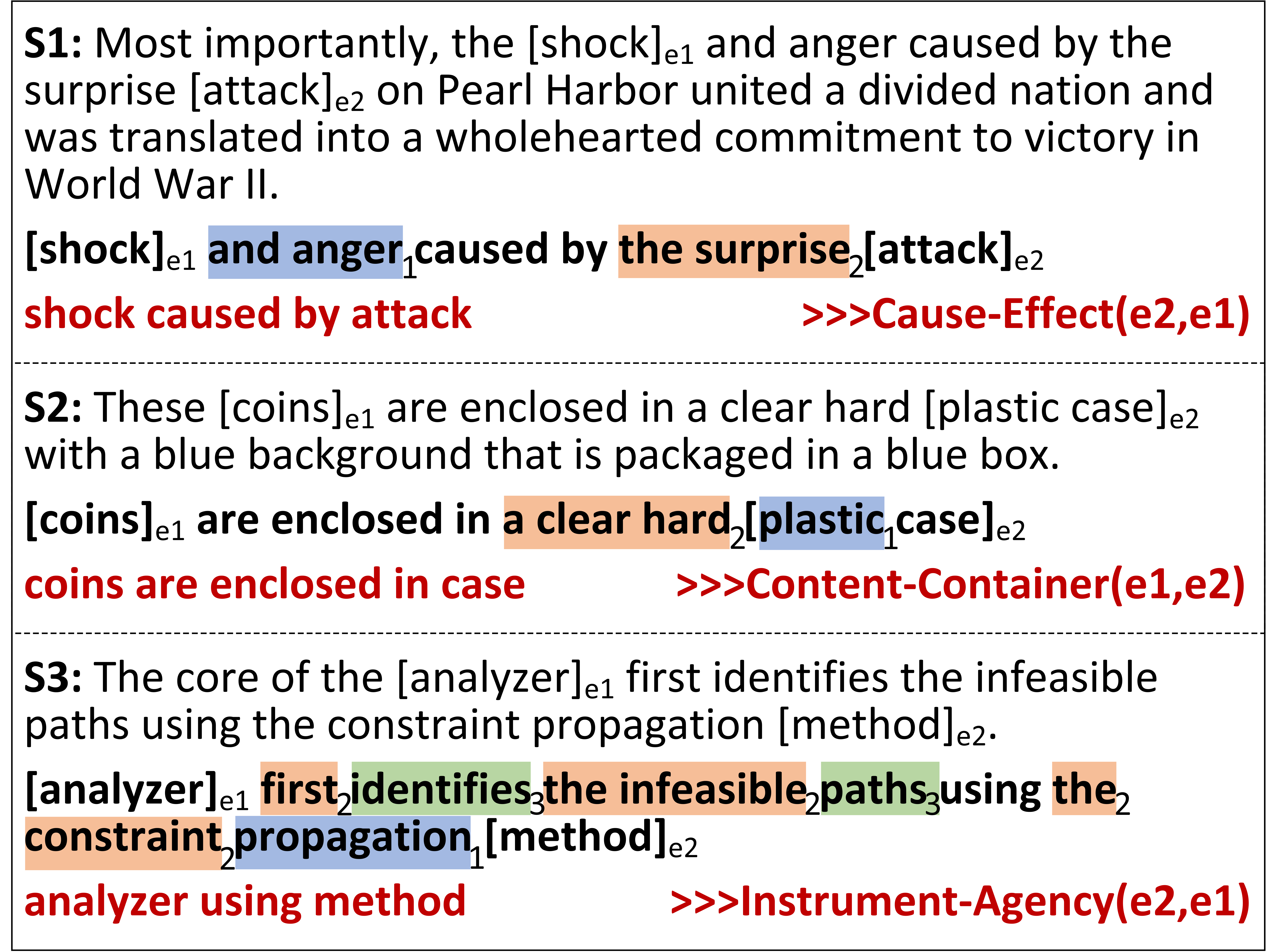}}
\caption{Syntactic Indicator Extraction. The blue highlights with a subscript 1 are removed abiding the first rule, {\em Entity Disambiguation}. And the orange highlights with a subscript 2 and the green highlights with a subscript 3 are removed respectively abiding {\em Principal Component Extraction} and {\em Unrelated Entities Removal}.}
\label{fig:3examples}
\end{figure}

We extract syntactic indicators from the text between two target entities by removing irrelevant words. Fortunately, the target subsequence is accessible from a sentence via entity markers. After that, we acquire the syntactic indicators under the guidance of syntactic knowledge, which can be characterized as follows:

\paragraph{Entity Disambiguation}
Nouns that are around with a conjunction word “and” or “or”, and compound nouns that consist of no less than two nouns will be disambiguated by removing the restrictive and supplementary words. As shown in Fig.~\ref{fig:3examples}, “shock and anger” is transformed to “shock”, “plastic case” and “propagation method” are transformed to “case” and “method”, we remove the highlighted parts marked with a subscript 1. Each instance in the labeled data contains only one relationship, like the relation in the first example of Fig.~\ref{fig:3examples} is about “shock” and “attack”, so nouns in target entities naturally remain.

\paragraph{Principal Component Extraction} 
Remove adjectives, adverbs and other modifiers from the text to obtain the principal components, expressing the primary semantic relations. In Fig.~\ref{fig:3examples}, the highlighted parts marked with a subscript 2 are removed from subsequences, such as [the, surprise], [a, clear, hard], and [first, the, infeasible, the, constraint].

\paragraph{Unrelated Entity Removal}
Remove any other named entity and the corresponding actions except two target entities to obtain an indicator sequence shaped like “{\em shock caused by attack}”, “{\em coins are enclosed in case}” and “{\em analyzer using method}” shown in Fig.~\ref{fig:3examples}. In the third instance, the irrelevant entity “paths” and its corresponding action “identifies” are removed. 

Finally, we acquire an exclusive indicator sequence from a given sentence, which deemed without any irrelevant words. The syntactic indicator is included between two target entities. 

\subsection{BERT-based Contextual Encoder}

The pre-trained BERT language representation model \cite{devlin2019bert} is a multi-layer bidirectional transformer encoder \cite{vaswani2017attention}, designed to pre-train deep bidirectional representations by jointly conditioning on both left and right context in all layers. The input of BERT can be able to a single sentence or a pair of sentences. A special token “[CLS]” is always the first token of each sequence. Sentence pairs are separated with a token “[SEP]” and packed together into a single sequence. BERT is the first fine-tuning based representation model for a wide range of tasks, such as question answering and language inference, without substantial task-specific architecture modifications. Because of the ubiquitous use of BERT recently, we will omit an exhaustive background description of the architecture of BERT. 

\paragraph{BERT Module}

Given the sentence \begin{math} S \end{math}, we insert four markers “e11”, “e12”, “e21” and “e22” at the beginning and end of two target entities (\begin{math} e_1 \end{math}, \begin{math} e_2 \end{math}), which conduces to capture the entity locations. While the corresponding indicator sequence \begin{math} S^* \end{math} always start with entity \begin{math} e_1 \end{math} and end with \begin{math} e_2 \end{math}, we insert “\begin{math}\#\end{math}” behind \begin{math} e_1 \end{math} and insert “\begin{math}\$\end{math}” before \begin{math} e_2 \end{math} to mark the syntactic indicator. 

To fine-tune BERT, we feed both two sequences into the WordPiece tokenizer and then concatenate the obtained sub-tokens into a single token sequence \begin{math} T \end{math}. Following the original implementation of BERT, we add a token [CLS] to the beginning of the token sequence and separate two sequences with a token [SEP]. Then, we feed \begin{math} T \end{math} into the BERT to produce the current representation of each token.

\paragraph{Aggregate Sequence Representation}

The final hidden state sequence \begin{math} H \end{math} output from the BERT module corresponds to the task-oriented embedding of each token. Suppose \begin{math}H_0\end{math} is the hidden state of first special token [CLS], we add an activation operation and a fully connected layer to obtain a vector \begin{math} H_{0}^{\prime} \end{math} as the representation of the aggregate sequence.
\begin{equation}
H_{0}^{\prime}=W_{0}\left(\tanh \left(H_{0}\right)\right)+b_{0}
\end{equation}

\paragraph{Entity Representations}

Hidden state \begin{math}H_{m}\end{math}, \begin{math}H_{n}\end{math}, \begin{math}H_{p}\end{math} and \begin{math}H_{q}\end{math} are vector representations of four entity markers “e11”, “e12”, “e21” and “e22”. For the target entities, vectors between \begin{math}H_{m}\end{math} and \begin{math}H_{n}\end{math} represent entity \begin{math} e_1 \end{math}, and vectors between \begin{math}H_{p}\end{math} and \begin{math}H_{q} \end{math} represent entity \begin{math} e_2 \end{math}. We apply an average operation to get a single vector representation following with a tanh activation operation and a fully connected layer. In this step, two entities share the same parameters \begin{math} W^{e} \end{math} and \begin{math} b^{e} \end{math}. As the following equations, the final representations of two target entities are respectively \begin{math} H^{e_1} \end{math},  \begin{math} H^{e_2}  \end{math}:
\begin{equation}
\begin{aligned}
H^{e_1} =&W^{e}\left[\tanh \left(\frac{1}{n-m+1} \sum_{t=m}^{n} H_{t}\right)\right]+b^{e}\\
H^{e_2} =&W^{e}\left[\tanh \left(\frac{1}{q-p+1} \sum_{t=p}^{q} H_{t}\right)\right]+b^{e}
\end{aligned}
\end{equation}

\paragraph{Syntactic Indicator Representation}

\begin{math}H_{i+1}\end{math} to \begin{math}H_{i+j}\end{math} are the hidden state vectors corresponds to the indicator sequence \begin{math} S^* \end{math}. We also apply an average operation following with a tanh activation operation and a fully connected layer to obtain the final representation:
\begin{equation}
z=W_{z}\left[\tanh \left(\frac{1}{j} \sum_{t=1}^{j} H_{i+t}\right)\right]+b_{z}
\end{equation}
where \begin{math} H_{i+t} \end{math} is the \begin{math} t^{th} \end{math} vector representation in \begin{math} S^* \end{math}. 

For fine-tuning, we concatenate \begin{math}H_{0}^{\prime}\end{math}, \begin{math}H^{e_1}\end{math}, \begin{math}H^{e_2}\end{math} , and \begin{math} z \end{math}, then consecutively add two fully connected layers with weights \begin{math} W_{1} \end{math}, \begin{math} W_{2} \end{math} and biases \begin{math} b_{1} \end{math}, \begin{math} b_{2} \end{math}. Finally, we obtain a relation representation vector \begin{math} r \end{math} used for classifying relations.
\begin{equation}
r=W_{2}\left[W_{1}\left[\text{concat} \left(H_{0}^{\prime}, H^{e_1}, H^{e_2}, z\right)\right]+b_{1}\right]+b_{2}
\end{equation}

\subsection{Relation classifer}

Given an instance \begin{math} x \end{math} with entire sentence \begin{math} S \end{math} and indicator sequence \begin{math} S^* \end{math}, we can obtaine the relation representation \begin{math} r \end{math} by the relation encoder. For classifying, we apply a fully connected softmax layer to produce a probability distribution \begin{math} p\left(y | x, \theta\right) \end{math} over all predefined relation types:
\begin{equation}
p\left(y | x, \theta\right) =\text{softmax}\left(W^{*}r+b^{*}\right)
\end{equation}
where \begin{math} y \in Y \end{math} is the target relation type, \begin{math} \theta \end{math} refers all learnable parameters in the network including \begin{math} W^{*} \in \mathbb{R}^{|Y| \times d_{h}} \end{math} and \begin{math} b^{*} \in \mathbb{R}^{|Y|} \end{math}, where \begin{math} |Y| \end{math} is the number of relation types. 

\subsection{Training Procedure}
For the perpose of making a clear distinction between different relation categories and reducing the influence of noise, we design our loss function based on the commonly used cross-entropy, referring to the rank loss function proposed by Santos et al. \cite{dos2015classifying}. The total loss \begin{math} \mathcal{L} \end{math} on a batch with the size of \begin{math} k \end{math} can be expressed as the following equation:
\begin{equation}
\begin{aligned}
\mathcal{L} =&-\sum_{i=1}^{k}\log p\left(y ^{+} | c, \theta\right)\\
 &-\beta \sum_{i=1}^{k} \log \left(1-p\left(y ^{-} | c, \theta\right)\right) + \lambda \| \theta \|_{2}^{2}
\end{aligned}
\end{equation}
where the first term in the right side decreases as the probability \begin{math} p\left(y ^{+} | c, \theta\right) \end{math} increases and the second term with a hyper-parameter \begin{math} \beta \end{math} in the right side decreases as the the probability \begin{math} p\left(y ^{-} | c, \theta\right) \end{math} decreases. For each instance, \begin{math} y ^{+} \in Y \end{math} is the correct relation label, while \begin{math} y ^{-} \in Y \end{math} is a negative category chose with the highest probability among all incorrect relation types in each training round:
\begin{equation}
y^{-}=\underset{y \in Y ; y \neq y^{+}}{\arg \max} p\left(y | x, \theta\right)
\end{equation}

In relation extraction, an artificial class {\em Other} is used to refer to the relation between target entities that does not belong to any of natural classes. Therefore, the class {\em Other} is too noisy to have common representative characteristics since it consists of many different categories of relations. For this reason, we calculate loss on each relation class except {\em Other} to reduce the impact of noise, reflected in the loss function as \begin{math} y ^{+} \neq Other \end{math} and \begin{math} y ^{-} \neq Other \end{math}.

To alleviate overfitting, we add a dropout layer before the fully connected softmax layer in training procedure and constrain the L2 regularization with a coefficient \begin{math} \lambda \end{math} as the third term in the right side.

\section{Experiments}

\subsection{Dataset and Evaluation Metric}
 
To evaluate the performance of our model, we conduct experiments on the SemEval-2010 Task 8 dataset \cite{hendrickx2009semeval}, the published benchmark for relation extraction. The dataset contains 10, 717 annotated instances, including 8, 000 instances for training and 2, 717 instances for testing. All instances are annotated with 9 directed relations types and an artificial class {\em Other}. Nine directed relations are respectively {\em Cause-Effect}, {\em Instrument-Agency}, {\em Product-Producer}, {\em Content-Container}, {\em Entity-Origin}, {\em Entity-Destination}, {\em Component-Whole}, {\em Member-Collection}, and {\em Message-Topic}. We take direction into consideration and the total number of relation types is 19. We adopt macro-averaged F1-score for nine actual relations (excluding {\em Other}) to evaluate the model, which is the official evaluation metric for SemEval-2010 Task 8.

\subsection{Experimental Settings}

\begin{table}[ht]
\renewcommand\arraystretch{1.5}
\caption{Hyper-parameters}
\begin{center}
\begin{tabular}{|l|c|}
\hline
\multicolumn{1}{|c|}{\textbf{Description}} &\multicolumn{1}{|c|}{\textbf {Value} }\\ \hline
Max Sequence Length after Tokenization & 128   \\ \hline
Batch Size for Training & 16    \\ \hline 
Initial Learning Rate for Adam                   & \begin{math} 2 \times 10^{-5} \end{math}  \\ \hline
Number of Training Epochs       & 5.0    \\ \hline
Dropout Rate     & 0.1    \\ \hline
L2 Regularixation Coefficient    & \begin{math} 5 \times 10^{-3} \end{math}  \\ \hline
Hyper-parameter \begin{math} \beta \end{math} in Loss Function &5.0\\ \hline
\end{tabular}
\label{table:table0}
\end{center}
\end{table}

For the pre-trained BERT model, we use the uncased model to integrate our approach. The hyper-parameters we set in the proposed model are shown in table~\ref{table:table0}. Furthermore, the parameters of the pre-trained BERT model are initialized according to the original \cite{devlin2019bert}.

\subsection{Result}

Results of various neural models are demonstrated in table~\ref{table:table1}. We achieve a strong empirical result based on the proposed approach.

\begin{table}[ht]
\renewcommand\arraystretch{1.5}
\caption{Performance comparison on extracting relations}
\begin{center}
\begin{tabular}{|l|c|}
\hline
\multicolumn{1}{|c|}{\textbf{Model}}    & \multicolumn{1}{|c|}{\textbf{F1}}    \\ \hline
 CNN (Zeng et al., 2014) \cite{zeng2014relation}                             & 78.9           \\
                                                             + WN                     & 82.7           \\ \hline 
CR-CNN (Santos et al., 2015)\cite{dos2015classifying}                                     & 84.1           \\ \hline 
 Attention CNN (Shen and Huang, 2016) \cite{huang2016attention}                     & 84.3       \\
                                            + POS, WN, WAN                             & 85.9           \\ \hline
 Bi-LSTM (Zhang et al., 2015) \cite{zhang2015relation}                                       & 82.7           \\
                                                  + POS, NER, DEP, WN            & 84.3           \\ \hline
 Attention Bi-LSTM (Zhou et al., 2016) \cite{zhou2016attention}                         & 84.0           \\ \hline
Hier Attention Bi-LSTM (Xiao and Liu, 2016) \cite{xiao2016semantic}                       & 84.3           \\ \hline 
Attention Bi-LSTM (Lee et al., 2019) \cite{lee2019semantic}                     & 84.7           \\
                                + LET                                      & 85.2           \\ \hline
 R-BERT (Wu et al., 2019) \cite{wu2019enriching}                                              & 89.25          \\ \hline 
 \textbf{Indicator-aware BERT (Ours)}                                   & \textbf{90.36} \\ \hline
\end{tabular}
\label{table:table1}
\end{center}
\end{table}

Table~\ref{table:table1} shows that our model obtains an F1-score of 90.36\%, outperforming the state-of-the-art models substantially. The best results of the CNN-based and RNN-based models range from 84\% to 86\%, while the recent R-BERT model proposed by Wu and He \cite{wu2019enriching} obtains the best F1-score of 89.25\%, which has an approximately 4-point gap with previous methods. It is noteworthy that the proposed relation extraction model introducing syntactic indicators has a further performance improvement in this task.

\subsection{Analysis}

To demonstrate that introducing syntactic indicators indeed affects relation extraction, we create two more settings to conduct experiments for comparison and further build another neural model without BERT structure for more forceful evidence. Experimental results shown in table~\ref{table:table3} provide ample proof that incorporating syntactic indicators indeed improves the performance of relation extraction. 

\begin{table}[ht]
\renewcommand\arraystretch{1.5}
\caption{Experimental results based on different input and models}
\begin{center}
\begin{tabular}{|c|l|l|r|}
\hline
\multicolumn{1}{|c|}{\textbf{Model}}     &\multicolumn{2}{|c|}{\textbf{Input}}     &\multicolumn{1}{|c|}{\textbf{F1}}   \\ \hline
\multirow{3}*{\textbf{BERT-based}}   
&\multicolumn{2}{|l|}{Entire Sentence + Indicator Sequence}         & \textbf{90.36}   \\ \cline{2-4}
&\multicolumn{2}{|l|}{Entire Sentence}         & 89.30                \\ \cline{2-4}
&\multicolumn{2}{|l|}{Indicator Sequence}    & 86.79                \\ \hline
\multirow{5}*{\textbf{Non-BERT}}  
&\multicolumn{1}{|c|}{\textbf{LSTM}}&\multicolumn{1}{|c|}{\textbf{CNN}}&\\ \cline{2-4}
&Entire Sentence & Indicator Sequence  &\textbf{85.9}  \\ \cline{2-4}
&Entire Sentence   &-                                 & 84.4   \\ \cline{2-4}
&-& Indicator Sequence                             & 82.5         \\ \cline{2-4}
&Entire Sentence&Entire Sentence            & 84.0          \\ \hline
\end{tabular}
\label{table:table3}
\end{center}
\end{table}

\paragraph{Experiments on BERT-based Model}
\begin{itemize}
\item  Two additional experiments only use one of the sequences and the experimental results are listed in lines two through four in table~\ref{table:table3}. The experiment only using the entire sentence as input produces an F1-score of 89.30\%, which is 1.06\% lower than the proposed approach. Although an indicator sequence just composed of a few words, the experiment only using the indicator sequence produces an F1-score of 86.79\%. It can be said that indicator sequences contain enough information for classifying relations but are likely to provide incomplete information. The proposed BERT-based model leverages both the syntactic indicator and the sentential context for relation extraction, which can be considered to be able to maintain a balance between reducing noise and capturing complete features.
\end{itemize}

\paragraph{Experiments on Non-BERT Model}
\begin{itemize}
\item We construct a model without BERT structure for further confirmations, which consists of a CNN module to capture the indicative features from indicator sequences and a Bi-LSTM module to capture the contextual information from entire sentences. Experimental results obtained from the Non-BERT structure model are listed in lines six through nine in table~\ref{table:table3}. The model obtains an F1-score of 85.9\% by combining the information from two modules, which outperforms the best CNN-based and RNN-based models. Even compared with the approaches using high-level lexical features such as WordNet, DPT, DEP, NLP tags or NER tags, it also has the best result. Likewise, we separately feed one of the sequences into the model. Correspondingly, the entire sentence is encoded using the Bi-LSTM module while the syntactic indicator is encoded using the CNN module. Unsurprisingly, both of the F1-scores are not bad but lower, which further proves the validity of the constraint on relation representations by syntactic indicators.
\item We further capture the features of the entire sentence twice using the CNN module and Bi-LSTM module respectively and then combine them to make a final prediction, the result becomes worse instead. This proves that noisy information unrelated to entity relations exist in the sentence, and excessive use of irrelevant features as relational features will degrade the performance of relation extraction. Therefore, it is very necessary to impose constraints on semantic relation representations to avoid the impact of noisy information.
\end{itemize}

\paragraph{Contributions of Syntactic Indicators}
\begin{itemize}
\item Table~\ref{table:table4} shows the contributions of syntactic indicators on precision, recall and F1-score for each relation category(performed on BERT-based model). The proposed model incorporating syntactic indicators increases the F1-score on each category, where the precisions on all categories except {\em Entity-Destination} are increased and the recalls on most categories are improved or remained the same. Especially, the precisions on {\em Instrument-Agency}, {\em Member-Collection}, {\em Message-Topic} and {\em Product-Producer} increased by 3.13, 3.06, 2.61 and 5.95 percentage points respectively. The effects of syntactic indicators are more prominently reflected on these categories because of instances containing such types of relations often have more noisy words in the text between target entities.
\end{itemize}

\begin{table}[t]
\renewcommand\arraystretch{1.5}
\caption{Contributions of Syntactic Indicators on Precision, Recall and F1-score for each Relation Category}
\begin{center}
\begin{tabular}{|c|c|c|c|c|c|c|}
\hline
\multirow{2}*{\textbf{Relation}} 
&\multicolumn{2}{|c|}{\textbf{Precision}} & \multicolumn{2}{|c|}{\textbf{Recall}} &\multicolumn{2}{|c|}{\textbf{F1-score}}    \\ \cline{2-7}
&-          &\textbf{+IS}                   &-           &\textbf{+IS}             &-         &\textbf{+IS}    \\ \hline 
\multicolumn{1}{|l|}{Cause-Effect}             
&93.27   &\textbf{94.48}                              &92.99    & \textbf{93.90}                      &93.13   & \textbf{94.19}    \\ \hline 
\multicolumn{1}{|l|}{Component-Whole}  
&86.52   &\textbf{88.46}                              &88.46   &88.46                        &87.48   &\textbf{88.46}     \\ \hline 
\multicolumn{1}{|l|}{Content-Container}   
&89.05    &\textbf{90.77}                             &\textbf{93.23}   &92.19                        &91.09   &\textbf{91.47}     \\ \hline 
\multicolumn{1}{|l|}{Entity-Destination}    
&\textbf{93.84}     &93.33                            &93.84       &\textbf{95.89}                      &93.84   &\textbf{94.59}  \\ \hline 
\multicolumn{1}{|l|}{Entity-Origin}            
  &89.66        &\textbf{90.77}                        &90.70     &\textbf{91.47}                       &90.17  &\textbf{91.12}    \\ \hline 
\multicolumn{1}{|l|}{Instrument-Agency}  
&85.92        &\textbf{89.05}                           &78.21     &78.21                   &81.88  &\textbf{83.28}     \\ \hline 
\multicolumn{1}{|l|}{Member-Collection}  
&84.49       &\textbf{87.55}                           &\textbf{88.84}    &87.55                      &86.61  &\textbf{87.55}   \\ \hline 
\multicolumn{1}{|l|}{Message-Topic}       
  &87.54	 &\textbf{90.15}                           &\textbf{96.93}     &94.64                     &92.00  &\textbf{92.34}   \\ \hline 
\multicolumn{1}{|l|}{Product-Producer}
 &84.84       &\textbf{90.79}                         &89.61     &89.61                    &87.16 &\textbf{90.20}   \\ \hline 
\multicolumn{2}{l}{IS: Indicator Sequence}
\end{tabular}
\label{table:table4}
\end{center}
\end{table}

\section{Conclusions}

In this paper, we propose syntactic indicators that are insensitive to lexical word forms and a novel indicator-aware neural model leveraging both syntactic indicators and sentential contexts to fulfill the relation extraction. The proposed approach performed on BERT-based model achieves an F1-score of 90.36\% in SemEval-2010 Task 8, outperforming the state-of-the-art methods. The implementation with the non-BERT model also achieves the best result in CNN-based and RNN-based models. Thanks to the incorporating of syntactic indicators, capturing more determinative features for classifying relations while reducing noise impact, our approach effectively improves the performance of relation extraction. 

In the future, we expect to leverage the syntactic indicators into more complex multi-relation extraction and distantly supervised relation extraction. Furthermore, we will research how to utilize the deep neural network to automatically locate the indicator in sentences, rather than extract indicators under the guidance of syntactic knowledge.

\section*{Acknowledgment}

The research reported in this paper was supported in part by the National Natural Science Foundation of China under the grant No. 91746203. 

\bibliographystyle{IEEEtran}
\bibliography{IEEEabrv, reference}

\begin{thebibliography}{10}
\providecommand{\url}[1]{#1}
\csname url@samestyle\endcsname
\providecommand{\newblock}{\relax}
\providecommand{\bibinfo}[2]{#2}
\providecommand{\BIBentrySTDinterwordspacing}{\spaceskip=0pt\relax}
\providecommand{\BIBentryALTinterwordstretchfactor}{4}
\providecommand{\BIBentryALTinterwordspacing}{\spaceskip=\fontdimen2\font plus
\BIBentryALTinterwordstretchfactor\fontdimen3\font minus
  \fontdimen4\font\relax}
\providecommand{\BIBforeignlanguage}[2]{{%
\expandafter\ifx\csname l@#1\endcsname\relax
\typeout{** WARNING: IEEEtran.bst: No hyphenation pattern has been}%
\typeout{** loaded for the language `#1'. Using the pattern for}%
\typeout{** the default language instead.}%
\else
\language=\csname l@#1\endcsname
\fi
#2}}
\providecommand{\BIBdecl}{\relax}
\BIBdecl

\bibitem{fader2011identifying}
A.~Fader, S.~Soderland, and O.~Etzioni, ``Identifying relations for open
  information extraction,'' in \emph{Proceedings of the conference on empirical
  methods in natural language processing}.\hskip 1em plus 0.5em minus
  0.4em\relax Association for Computational Linguistics, 2011, pp. 1535--1545.

\bibitem{wu2010open}
F.~Wu and D.~S. Weld, ``Open information extraction using wikipedia,'' in
  \emph{Proceedings of the 48th annual meeting of the association for
  computational linguistics}.\hskip 1em plus 0.5em minus 0.4em\relax
  Association for Computational Linguistics, 2010, pp. 118--127.

\bibitem{fan2005using}
R.~S. J. J.~Y. Fan, T.~H. C. T.-S. Chua, and M.-Y. Kan, ``Using syntactic and
  semantic relation analysis in question answering,'' in \emph{Proceedings of
  the 14th Text REtrieval Conference (TREC)}, vol. Special Publication
  500-266.\hskip 1em plus 0.5em minus 0.4em\relax National Institute of
  Standards and Technology, 2005.

\bibitem{yih2014semantic}
W.-t. Yih, X.~He, and C.~Meek, ``Semantic parsing for single-relation question
  answering,'' in \emph{Proceedings of the 52nd Annual Meeting of the
  Association for Computational Linguistics (Volume 2: Short Papers)}.\hskip
  1em plus 0.5em minus 0.4em\relax Association for Computational Linguistics,
  2014, pp. 643--648.

\bibitem{miller1991semantic}
G.~A. Miller and C.~Fellbaum, ``Semantic networks of english,''
  \emph{Cognition}, vol.~41, no. 1-3, pp. 197--229, 1991.

\bibitem{vossen1998multilingual}
P.~Vossen, ``A multilingual database with lexical semantic networks,''
  \emph{Dordrecht: Kluwer Academic Publishers. doi}, vol.~10, pp. 978--94,
  1998.

\bibitem{zeng2014relation}
D.~Zeng, K.~Liu, S.~Lai, G.~Zhou, and J.~Zhao, ``Relation classification via
  convolutional deep neural network,'' in \emph{Proceedings of COLING 2014, the
  25th International Conference on Computational Linguistics: Technical
  Papers}.\hskip 1em plus 0.5em minus 0.4em\relax Association for Computational
  Linguistics, 2014, pp. 2335--2344.

\bibitem{dos2015classifying}
C.~dos Santos, B.~Xiang, and B.~Zhou, ``Classifying relations by ranking with
  convolutional neural networks,'' in \emph{Proceedings of the 53rd Annual
  Meeting of the Association for Computational Linguistics and the 7th
  International Joint Conference on Natural Language Processing (Volume 1: Long
  Papers)}.\hskip 1em plus 0.5em minus 0.4em\relax Association for
  Computational Linguistics, 2015, pp. 626--634.

\bibitem{zhang2015bidirectional}
S.~Zhang, D.~Zheng, X.~Hu, and M.~Yang, ``Bidirectional long short-term memory
  networks for relation classification,'' in \emph{Proceedings of the 29th
  Pacific Asia conference on language, information and computation}.\hskip 1em
  plus 0.5em minus 0.4em\relax Association for Computational Linguistics, 2015,
  pp. 73--78.

\bibitem{zhou2016attention}
P.~Zhou, W.~Shi, J.~Tian, Z.~Qi, B.~Li, H.~Hao, and B.~Xu, ``Attention-based
  bidirectional long short-term memory networks for relation classification,''
  in \emph{Proceedings of the 54th Annual Meeting of the Association for
  Computational Linguistics (Volume 2: Short Papers)}, vol.~2.\hskip 1em plus
  0.5em minus 0.4em\relax Association for Computational Linguistics, 2016, pp.
  207--212.

\bibitem{xiao2016semantic}
M.~Xiao and C.~Liu, ``Semantic relation classification via hierarchical
  recurrent neural network with attention,'' in \emph{Proceedings of COLING
  2016, the 26th International Conference on Computational Linguistics:
  Technical Papers}.\hskip 1em plus 0.5em minus 0.4em\relax Association for
  Computational Linguistics, 2016, pp. 1254--1263.

\bibitem{huang2016attention}
X.~Huang \emph{et~al.}, ``Attention-based convolutional neural network for
  semantic relation extraction,'' in \emph{Proceedings of COLING 2016, the 26th
  International Conference on Computational Linguistics: Technical
  Papers}.\hskip 1em plus 0.5em minus 0.4em\relax Association for Computational
  Linguistics, 2016, pp. 2526--2536.

\bibitem{lee2019semantic}
J.~Lee, S.~Seo, and Y.~S. Choi, ``Semantic relation classification via
  bidirectional lstm networks with entity-aware attention using latent entity
  typing,'' \emph{Symmetry}, vol.~11, no.~6, p. 785, 2019.

\bibitem{fader2014open}
A.~Fader, L.~Zettlemoyer, and O.~Etzioni, ``Open question answering over
  curated and extracted knowledge bases,'' in \emph{Proceedings of the 20th ACM
  SIGKDD international conference on Knowledge discovery and data
  mining}.\hskip 1em plus 0.5em minus 0.4em\relax ACM, 2014, pp. 1156--1165.

\bibitem{bjorne2011extracting}
J.~Bj{\"o}rne, J.~Heimonen, F.~Ginter, A.~Airola, T.~Pahikkala, and
  T.~Salakoski, ``Extracting contextualized complex biological events with rich
  graph-based feature sets,'' \emph{Computational Intelligence}, vol.~27,
  no.~4, pp. 541--557, 2011.

\bibitem{devlin2019bert}
J.~Devlin, M.-W. Chang, K.~Lee, and K.~Toutanova, ``Bert: Pre-training of deep
  bidirectional transformers for language understanding,'' in \emph{Proceedings
  of the 2019 Conference of the North American Chapter of the Association for
  Computational Linguistics: Human Language Technologies, Volume 1 (Long and
  Short Papers)}.\hskip 1em plus 0.5em minus 0.4em\relax Association for
  Computational Linguistics, 2019, pp. 4171--4186.

\bibitem{kambhatla2004combining}
N.~Kambhatla, ``Combining lexical, syntactic, and semantic features with
  maximum entropy models for extracting relations,'' in \emph{Proceedings of
  the ACL 2004 on Interactive poster and demonstration sessions}.\hskip 1em
  plus 0.5em minus 0.4em\relax Association for Computational Linguistics, 2004,
  p.~22.

\bibitem{suchanek2006combining}
F.~M. Suchanek, G.~Ifrim, and G.~Weikum, ``Combining linguistic and statistical
  analysis to extract relations from web documents,'' in \emph{Proceedings of
  the 12th ACM SIGKDD international conference on Knowledge discovery and data
  mining}.\hskip 1em plus 0.5em minus 0.4em\relax ACM, 2006, pp. 712--717.

\bibitem{qian2008exploiting}
L.~Qian, G.~Zhou, F.~Kong, Q.~Zhu, and P.~Qian, ``Exploiting constituent
  dependencies for tree kernel-based semantic relation extraction,'' in
  \emph{Proceedings of the 22nd International Conference on Computational
  Linguistics-Volume 1}.\hskip 1em plus 0.5em minus 0.4em\relax Association for
  Computational Linguistics, 2008, pp. 697--704.

\bibitem{mooney2006subsequence}
R.~J. Mooney and R.~C. Bunescu, ``Subsequence kernels for relation
  extraction,'' in \emph{Advances in neural information processing systems},
  2006, pp. 171--178.

\bibitem{zhang2015relation}
D.~Zhang and D.~Wang, ``Relation classification via recurrent neural network,''
  \emph{arXiv preprint arXiv:1508.01006}, 2015.

\bibitem{dai2015semi}
A.~M. Dai and Q.~V. Le, ``Semi-supervised sequence learning,'' in
  \emph{Advances in neural information processing systems}, 2015, pp.
  3079--3087.

\bibitem{howard2018universal}
J.~Howard and S.~Ruder, ``Universal language model fine-tuning for text
  classification,'' in \emph{Proceedings of the 56th Annual Meeting of the
  Association for Computational Linguistics (Volume 1: Long Papers)}.\hskip 1em
  plus 0.5em minus 0.4em\relax Association for Computational Linguistics, 2018,
  pp. 328--339.

\bibitem{wu2019enriching}
S.~Wu and Y.~He, ``Enriching pre-trained language model with entity information
  for relation classification,'' \emph{arXiv preprint arXiv:1905.08284}, 2019.

\bibitem{socher2012semantic}
R.~Socher, B.~Huval, C.~D. Manning, and A.~Y. Ng, ``Semantic compositionality
  through recursive matrix-vector spaces,'' in \emph{Proceedings of the 2012
  joint conference on empirical methods in natural language processing and
  computational natural language learning}.\hskip 1em plus 0.5em minus
  0.4em\relax Association for Computational Linguistics, 2012, pp. 1201--1211.

\bibitem{mintz2009distant}
M.~Mintz, S.~Bills, R.~Snow, and D.~Jurafsky, ``Distant supervision for
  relation extraction without labeled data,'' in \emph{Proceedings of the Joint
  Conference of the 47th Annual Meeting of the ACL and the 4th International
  Joint Conference on Natural Language Processing of the AFNLP}.\hskip 1em plus
  0.5em minus 0.4em\relax Association for Computational Linguistics, 2009, pp.
  1003--1011.

\bibitem{zeng2015distant}
D.~Zeng, K.~Liu, Y.~Chen, and J.~Zhao, ``Distant supervision for relation
  extraction via piecewise convolutional neural networks,'' in
  \emph{Proceedings of the 2015 Conference on Empirical Methods in Natural
  Language Processing}.\hskip 1em plus 0.5em minus 0.4em\relax The Association
  for Computational Linguistics, 2015, pp. 1753--1762.

\bibitem{min2013distant}
B.~Min, R.~Grishman, L.~Wan, C.~Wang, and D.~Gondek, ``Distant supervision for
  relation extraction with an incomplete knowledge base,'' in \emph{Proceedings
  of the 2013 Conference of the North American Chapter of the Association for
  Computational Linguistics: Human Language Technologies}.\hskip 1em plus 0.5em
  minus 0.4em\relax The Association for Computational Linguistics, 2013, pp.
  777--782.

\bibitem{sennrich2016neural}
R.~Sennrich, B.~Haddow, and A.~Birch, ``Neural machine translation of rare
  words with subword units,'' in \emph{Proceedings of the 54th Annual Meeting
  of the Association for Computational Linguistics (Volume 1: Long
  Papers)}.\hskip 1em plus 0.5em minus 0.4em\relax Association for
  Computational Linguistics, 2016, pp. 1715--1725.

\bibitem{vaswani2017attention}
A.~Vaswani, N.~Shazeer, N.~Parmar, J.~Uszkoreit, L.~Jones, A.~N. Gomez,
  {\L}.~Kaiser, and I.~Polosukhin, ``Attention is all you need,'' in
  \emph{Advances in neural information processing systems}, 2017, pp.
  5998--6008.

\bibitem{hendrickx2009semeval}
I.~Hendrickx, S.~N. Kim, Z.~Kozareva, P.~Nakov, D.~{\'O}~S{\'e}aghdha,
  S.~Pad{\'o}, M.~Pennacchiotti, L.~Romano, and S.~Szpakowicz, ``Semeval-2010
  task 8: Multi-way classification of semantic relations between pairs of
  nominals,'' in \emph{Proceedings of the Workshop on Semantic Evaluations:
  Recent Achievements and Future Directions}.\hskip 1em plus 0.5em minus
  0.4em\relax Association for Computational Linguistics, 2009, pp. 94--99.

\end{thebibliography}

\end{document}